\documentclass[conference]{IEEEtran}

\IEEEoverridecommandlockouts
\usepackage{cite}
\usepackage{amsmath,amssymb,amsfonts}
\usepackage{algorithmic}
\usepackage{graphicx}
\usepackage{textcomp}
\usepackage{xcolor}
\usepackage{float}
\usepackage{multirow}
\usepackage[hidelinks]{hyperref}
\usepackage{multirow}
\usepackage{array}
\usepackage{float}
\usepackage{color}
\usepackage[switch]{lineno}
\usepackage{bbding}

\usepackage{balance} 

\def\BibTeX{{\rm B\kern-.05em{\sc i\kern-.025em b}\kern-.08em
    T\kern-.1667em\lower.7ex\hbox{E}\kern-.125emX}}
\begin{document}

\title{MPRE: Multi-perspective Patient Representation Extractor for Disease Prediction\\}



\DeclareRobustCommand*{\IEEEauthorrefmark}[1]{%
	\raisebox{0pt}[0pt][0pt]{\textsuperscript{\footnotesize #1}}%
}
\author{
	\IEEEauthorblockN{
		Ziyue Yu\IEEEauthorrefmark{1},
		Jiayi Wang\IEEEauthorrefmark{1},
		Wuman Luo\IEEEauthorrefmark{1, 2$\ast$},
        Rita Tse\IEEEauthorrefmark{1, 2} and
		Giovanni Pau\IEEEauthorrefmark{3, 4}
	}
 \thanks{$\ast$ denotes the corresponding author}

	\IEEEauthorblockA{
		\IEEEauthorrefmark{1}Faculty of Applied Sciences, Macao Polytechnic University, China\\
		\IEEEauthorrefmark{2}Engineering Research Centre of Applied Technology on Machine Translation and\\ Artificial Intelligence of Ministry of Education, Macao Polytechnic University, China\\
		\IEEEauthorrefmark{3}Department of Computer Science and Engineering, University of Bologna, Italy\\
        \IEEEauthorrefmark{4}Department of Computer Science, University of California, Los Angeles, USA\\
        \{ziyue.yu, P1807511, luowuman, ritatse\}@mpu.edu.mo,
        giovanni.pau@unibo.it
    	}
}


\maketitle

\begin{abstract}
Patient representation learning based on electronic health records (EHR) is a critical task for disease prediction. This task aims to effectively extract useful information on dynamic features. Although various existing works have achieved remarkable progress, the model performance can be further improved by fully extracting the trends, variations, and the correlation between the trends and variations in dynamic features. In addition, sparse visit records limit the performance of deep learning models. To address these issues, we propose the Multi-perspective Patient Representation Extractor (MPRE) for disease prediction. Specifically, we propose Frequency Transformation Module (FTM) to extract the trend and variation information of dynamic features in the time-frequency domain, which can enhance the feature representation. In the 2D Multi-Extraction Network (2D MEN), we form the 2D temporal tensor based on trend and variation. Then, the correlations between trend and variation are captured by the proposed dilated operation. Moreover, we propose the First-Order Difference Attention Mechanism (FODAM) to calculate the contributions of differences in adjacent variations to the disease diagnosis adaptively. To evaluate the performance of MPRE and baseline methods, we conduct extensive experiments on two real-world public datasets. The experiment results show that MPRE outperforms state-of-the-art baseline methods in terms of AUROC and AUPRC.

\end{abstract}

\begin{IEEEkeywords}
Disease Prediction, Patient Representation, Visit Records
\end{IEEEkeywords}

\section{Introduction}
Patient representation aims at how to detect and represent useful information about each patient related to a medical diagnosis. Nowadays, patient representation based on electronic health records (EHR)~\cite{1.1, n1} is becoming increasingly important for disease prediction~\cite{yu2023dmnet}. Typically, EHR of a patient is a list of temporally ordered visit data. Each visit consists of three parts, i.e., static features (e.g., demographic information), dynamic features (e.g., medical lab test information), and diagnostic results~\cite{1.2, n8}. Static features are recorded only once at the patient's initial hospital visit, while dynamic features are updated at every subsequent visit. In this paper, we focus on how to effectively learn patient representation for disease prediction based on EHR. 

So far, various patient representation methods based on EHR have been proposed for disease prediction. Most of them are hybrids or variants of convolutional neural networks (CNN), recurrent neural networks (RNN), and attention mechanisms. Specifically, one-dimensional (1D) CNN~\cite{2.5}~\cite{n11} with different kernel sizes can extract the adjacent variations in visit records. Temporal convolutional network (TCN)~\cite{2.2} used 1D kernels to capture contextual information between visit records by crossing different time steps. To model visit records from multiple time scales, AdaCare was proposed to capture the long-term trend and short-term variation of biomarkers at different visit records. SAnD~\cite{99.4} adopted the Transformer-based framework with the masked self-attention mechanism for incorporating the information in time order. StageNet~\cite{4.5} proposed the stage-aware Long Short-term Memory (LSTM) to identify the stages (e.g., deterioration and recovery stages) of disease progression. These works have greatly improved the performance of patient representation, and have made considerable achievements in disease prediction. 

However, these methods still have much room for improvement in fully detecting the useful information hidden in dynamic features such as long-term and short-term trends, variations, and correlations between trends and variations~\cite{1.4, n7}. Usually, an upward trend in creatinine level indicates the patient is at risk for kidney disease~\cite{1.8}. The abnormal increase in bicarbonate levels indicates that the patient may have metabolic alkalosis~\cite{1.7}.  In addition, capturing the correlations between trends and variations of dynamic features is an important reference in medical diagnosis. For example, the positive correlation between the trend and variation in blood albumin levels indicates an upward trend with a gradual increasing pattern of variation, which causes acute inflammation to patients~\cite{n19}. 

Detecting hidden useful information of dynamic features is challenging due to data sparsity~\cite{miotto2018deep}. According to the statistics of modern popular EHR datasets~\cite{data1,data2,n20}, the average patient visit is only 10, the intervals between visits are irregular, and the average interval between two contiguous visits is as large as 2.5 months. As a result, the sparsity of patient visits limits the ability of the deep learning models to detect the hidden useful information of dynamic features.

To address the above issues, in this paper, we propose a framework called Multi-perspective Patient Representation Extractor (MPRE) for disease prediction based on EHR. The goal is to effectively detect the useful information (e.g., long-term and short-term trends, variations, and correlations between trends and variations) of dynamic features for patient representation. Specifically, we adopt symlets wavelet decomposition in Frequency Transformation Module (FTM) to capture the trend and variation information of dynamic features in the time-frequency domain. The reasons~\cite{n15, wang2018multilevel} why we use wavelet decomposition to detect trend and variation are: 1) it has the multi-scale analysis capability, which can obtain the representation of the visit records based on different time-frequency domains, 2) it possesses the property of time-frequency localization, which can effectively capture the rapid changes of dynamic features and the moment of peak appearance, and 3) dynamic features exhibit poor Cyclicality due to data sparsity, which makes traditional time series decomposition methods (e.g., seasonal and trend decomposition using loess) inapplicable. 

Besides, we design a 2D Multi-Extraction Network (2D MEN) to transform the 1D trend and variation information into the 2D temporal tensor. The goal is to capture the correlations between trends and variations. In addition, we propose the First-Order Difference Attention Mechanism (FODAM) to adaptively calculate the contribution of differences in adjacent variations of dynamic features. As pointed out in the American Diabetes Association~\cite{carnevale2003significance}, it is necessary to leverage the differences between adjacent variations in disease diagnosis. For instance, alternating positive and negative fluctuations in blood glucose levels often indicate abnormal insulin secretion. Additionally, differences in adjacent variations may have varying impacts on disease diagnosis~\cite{han2010chronic}.

In summary, the main contributions of this paper are listed as follows:

\begin{enumerate}
\item We propose MPRE to enhance the learning of patient representation for disease prediction based on EHR. MPRE consists of three major modules, namely FTM, 2D MEN, and FODAM. 

\item We propose FTM to extract the trend and variation from dynamic features. FTM can transform dynamic features into time-frequency domains to obtain more hidden information.

\item We propose 2D MEN to capture the correlation between the trend and variation. We propose FODAM to adaptively calculate the contribution of the differences in dynamic features' variations to disease diagnosis.

\item We implement the MPRE using the two real-world public datasets, namely “SCRIPT CarpeDiem Dataset” and “Health Facts Database”~\cite{data1,data2}. To evaluate the performance of the MPRE, we select the area under the receiver operating characteristic curve (AUROC) and precision-recall curve (AUPRC) as evaluation metrics. The experiment results show that the MPRE outperforms state-of-the-art baseline methods.
\end{enumerate}

The remainder of this paper is organized as follows. In Section II, we review related works for the disease prediction task. The methods of MPRE will be given in Section III. Section IV discusses the experiments of MPRE and baseline methods. We summarize this paper in Section V.

\section{Related Works}
In this section, we review the major works in disease prediction tasks from two perspectives, namely variation pattern detection methods and time-aware methods.\\
\textbf{Variation Pattern Detection Methods}.
Typically, 1D CNN-based methods used different kernel sizes to extract the local variation between the patients' visit records~\cite{2.5,n5,n11, n12}. Choi et al. proposed a reverse temporal attention mechanism with the RNN model called RETAIN to analyze the importance of visit records~\cite{99.1}. It can identify important clinical features and give higher attention weights to recent visit records. Ma et al. used bidirectional RNN with three attention mechanisms named Dipole to model visit records~\cite{99.3}. They believed that all historical visit records needed to be considered. Song et al. proposed a Transformer-based~\cite{2.4} model with a masked self-attention mechanism called SAnD. This work can aggregate the patients' information based on visit records in time order~\cite{99.4}.\\
\textbf{Time-aware Methods}.
Another perspective on the disease prediction task is the time-aware method. The temporal convolutional network (TCN) is the improved model based on 1D CNN. It proposed causal and dilation convolution to consider the temporal information between the adjacent patient visit records~\cite{2.1,2.2, 4.3, n9, n10}. Ma et al. proposed a multiple-scale dilation convolutional module called AdaCare to extract the different time scale information~\cite{1.3}. For the variants of RNN, T-LSTM was proposed to consider the patient information decreases with increasing time intervals~\cite{99.2}. It adopted the time-aware mechanism for extracting the variations of dynamic features. More recently, Gao et al. proposed StageNet to extract the stage of the disease progression~\cite{4.5}. This method adopted stage-aware long short-term memory to decide the time period with health status progression. ConCare adopted the time-aware attention mechanism and time information decay function to improve the model performance~\cite{n16}.

These works have achieved remarkable results in disease prediction, 
but we can further improve the performance of the model by fully capturing the trends, variations, and the correlation between trends and variations. In addition, sparse visit records and irregular visit intervals limit the performance of deep learning models.

\begin{figure*}[!t]
\centerline{\includegraphics[width=1\textwidth,height=0.46
\textwidth]{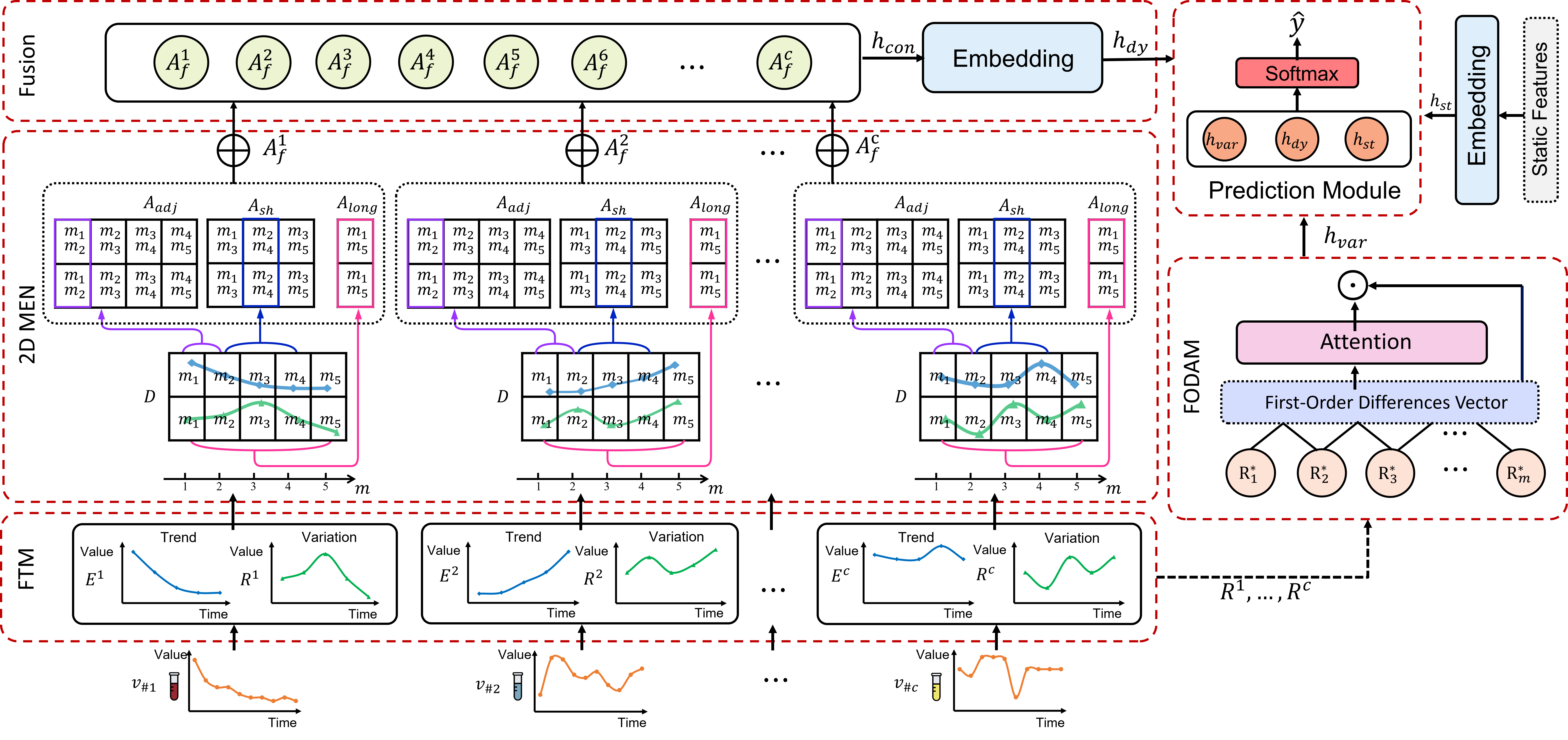}}
\caption{The overview of MPRE. Specifically, we adopt $\#$ as a specific dynamic feature across all visit records, and $*$ means the one specific dynamic feature. FTM is used to capture trend and variation information from the dynamic features separately. 2D MEN reshapes the trend and variation as the 2D temporal tensor and further captures the adjacent, short, and long-term correlation between the trend and variation. Fusion is used to embed all dynamic feature representations based on the outputs from 2D MEN. FODAM adaptively computes the contribution of variation differences. Static features are embedded by the linear layer. Finally, the prediction module aggregates the output of fusion, the result of FODAM, and the embedded static features to perform disease prediction.}
\label{overview}
\end{figure*}


\section{Methodology}
In this section, we first give the problem formulation of the disease prediction task. Then, we describe the overview of MPRE. After that, we introduce the major modules from the proposed MPRE in detail. Finally, we introduce the loss function and optimizer.

\subsection{Problem Formulation}
For temporal patient's visit records $V$, suppose that each patient has $t = (1, 2, \dots, T)$ visits with $|c|$ dynamic features (e.g., hemoglobin, creatinine). $V$ can be represented as follow:

\begin{equation}
\label{equation_inputx}
V  = \begin{pmatrix}  
  v_{11} & v_{12}  & \dots & v_{1,c}\\
  v_{21} & v_{22} & \dots& v_{2,c}\\
  \vdots &  &  &  \vdots\\
  v_{t1} & v_{t2}  & \dots& v_{t,c}
\end{pmatrix} \in \mathbb{R}^{t\times c},
\end{equation}
we denote $s \in \mathbb{R}^{s}$ as the vector of static features (e.g., gender, ethnicity).

In this paper, our predictive objective is presented as the multi-class disease prediction task. Specifically, given $V$ and $s$, each patient has a corresponding label $y \in \mathbb{R}^{d} (d = 0,1, ..., D)$, where $d$ means the number of types of the same disease, we aim to predict whether the patient will have the disease $y$ in the future. We formulate the objective as a prediction function $\hat{y}=\mathbb{F}(V,s; \Theta)$, where $\Theta$ means the model parameters and $\hat{y}$ is the predicted result. 

\subsection{Framework Overview}
Figure \ref{overview} shows the overview of MPRE. 
Initially, each dynamic feature is individually inputted into the FTM to obtain the corresponding time-frequency domain information (i.e., trend and variation information). Subsequently, the trend and variation are reshaped to form a 2D temporal tensor and use the proposed 2D MEN to capture the correlation between the trend and variation. In addition, 2D MEN can also extract the adjacent, short and long-term correlation to maintain the dependencies of trends and variations. We use the proposed FODAM to identify the contributions of the difference of adjacent variations for disease diagnosis. We adopt the linear layer to embed the static features. Fusion is used to concatenate the output of each 2D MEN and feed the concatenated result through the embedding layer. Finally, the embedded static features, the output of Fusion, and the results of FODAM are inputted into the prediction module to perform the disease prediction.

\subsection{Frequency Transformation Module}
In this paper, we adopt symlets wavelet to decompose each dynamic feature separately. The decomposed result reflects the time and frequency information of dynamic features. The low-frequency components can be obtained by low pass filter, and used to indicate the trend information. And the high-frequency components are obtained by high pass filter, which can be utilized to express the variation information. The filtered results enhance the representation of dynamic features. In addition, we adopt the symmetric mode to avoid boundary effects and to reduce the artifacts at the boundary~\cite{3.2}. The equations of FTM can be represented as follows:
\begin{equation}
    x = Column(V) \vspace{1.0ex} 
\end{equation}
\begin{equation}
E^{*} = \sum_{n=1}^{T} x_{n} \cdot \frac{1}{\sqrt{2}}\cdot \phi_{n} 
\end{equation}
\begin{equation}
    R^{*} = \sum_{n=0}^{T} x_{n} \cdot \frac{1}{\sqrt{2}}\cdot \psi_n
\end{equation}
where $Column(\cdot)$ is used to extract each column of $V$, i.e., the value vector $x \in \mathbb{R}^{t}$ for a specific dynamic feature across all visit records. $E = [E^{1}, E^{2}, \dots, E^{c}] \in \mathbb{R}^{c \times m}$ means the trend information. We adopt $ E^{*} \in \mathbb{R}^m$ as the trend information of one specific dynamic feature. $R = [R^{1}, R^{2}, \dots, R^{c}]\in \mathbb{R}^{c \times m}$ is the variation information. And $R^{*} \in \mathbb{R}^m$ denotes the variation information of one specific dynamic feature. $\phi_n$ and $\psi_n$ denote the low pass and high pass filter, respectively. In the symlets wavelet function, the length and coefficients of the filter depend on the selected type of symlets. 

\subsection{2D Multi-Extraction Network}

In 2D MEN, we aim to capture the correlation between the trend and variation. In this module, we first reshape the trend and variation information obtained from the FTM to form the 2D temporal tensor. The process of this operation can be represented as follows:
\begin{equation}
    D = Reshape(E^{*}; R^{*}) \vspace{1.0ex} 
\end{equation}
where $Reshape(\cdot)$ represents the dimension transformation operation. After that, we apply the proposed 2D MEN on 2D tensor $D \in \mathbb{R}^{2 \times m}$. The idea of 2D MEN is inspired by the work of Yu et al. and Ma et al~\cite{99.5,3.4}. These two works showed that local information can be captured in images and 1D temporal data by using dilated convolution. But different from these works, we achieve the dilated convolution with different trend and variation spans. Moreover, our dilated convolution is adapted to the 2D temporal tensor. Specifically, a standard CNN (i.e., dilation rate is 0) can capture adjacent correlations. A dilation rate of 1 captures short-term correlations, while a dilation rate greater than 2 extracts long-term correlations. In this paper, we adopt different dilation rates for each dynamic feature to obtain adjacent, short and long-term correlations simultaneously. Mathematically, the equation of the proposed dilated operation is shown as follows:

\begin{equation}
a_{p,q} = \sum_{k=1}^2 \sum_{l=1}^L d_{p, q+bl} \cdot G_{k,l} \vspace{1ex} 
\end{equation}
where $d$ is the value from 2D tensor $D$. $a_{p, q}$ is the output of dilated operation. $b$ denotes the dilation rate. $G_{k,l}$ means the kernel. $k$ and $l$ is the length. Based on the different dilation rates, we can obtain three feature maps, i.e., $A_{adj} \in \mathbb{R}^{q_{a}}, A_{sh} \in \mathbb{R}^{q_{s}}, A_{long} \in \mathbb{R}^{q_{l}}$, which represent the adjacent, short and long-term correlation between the trend and variation, respectively. 

The final step of 2D MEN is to concatenate the adjacent, short and long-term correlations to form the better dynamic features representation:
\begin{equation}
    A_{f} = Concat(A_{adj}; A_{sh}; A_{long}) \vspace{1.0ex} 
\end{equation}
where $Concat(\cdot)$ represents the concatenation operation. $A_{f} \in \mathbb{R}^{2 \times (q_{a}+q_{s}+q_{l})}$ represents the output of 2D MEN module. We adopt $[A_{f}^1, ..., A_{f}^c]$ as the representation of all dynamic features.

\subsection{First-Order Difference Attention Mechanism}
Based on the extracted variations from FTM, we propose FODAM to adaptively compute the contributions of differences in adjacent variations to the disease progression. The FODAM can be represented as follows:
\begin{equation}
\alpha = softmax(\frac{(R_{2:m}^{*} - R_{1:m-1}^{*}) (R_{2:m}^{*} - R_{1:m-1}^{*})^T}{\sqrt{dim}}) \vspace{1.0ex} 
\end{equation}
\begin{equation}
    h_{var} = \alpha \odot [R_{2:m}^{*} - R_{1:m-1}^{*}] \vspace{1.0ex} 
\end{equation}
where $\alpha \in \mathbb{R}^{m-1}$ denotes the attention weight of first-order differences. $R_{2:m}^{*} = [R_2^{*}, R_3^{*}, \ldots, R_m^{*}]$, $R_{1:m-1}^{*}=[R_1^{*}, R_2^{*}, \ldots, R_{m-1}^{*}]$, $R^{*}=[R_1^{*}, R_2^{*}, \ldots, R_m^{*}]$. $dim$ represents the dimension of $R^{*}$. We adopt the $\sqrt{dim}$ to avoid very small gradients in the training process~\cite{3.5}. $h_{var} \in \mathbb{R}^{m-1}$ is the representation of first-order variation difference. $\odot$ means the Hadamard product.

\subsection{Prediction Module}
The aim of the prediction module is to use the representation of static, dynamic features, and first-order variation differences to perform disease prediction. We first embed the static features based on the following equation:
\begin{equation}
    h_{st} = W_{s} s + b_{s} \vspace{1.0ex} 
\end{equation}
where $h_{st} \in \mathbb{R}^{d}$ is the representation of static features. $W_{s} \in \mathbb{R}^{d \times s}$ is learnable parameter, and $b_s \in \mathbb{R}^{d}$ is the bias.

Then, we do the fusion to embed all dynamic feature representations based on the 2D MEN.
\begin{equation}
    h_{con} = Concat(A_{f}^1; ...; A_{f}^c) 
\end{equation}
\begin{equation}
    h_{dy} = W_{d} h_{con} +b_{d} 
\end{equation}
where $h_{con} \in \mathbb{R}^{2 \times ((q_{a}+q_{s}+q_{l})*c)}$ means concatenating the results of each 2D MEN. $h_{dy} \in \mathbb{R}^{(q_{a}+q_{s}+q_{l})*c}$ denotes the representation of dynamic features. $W_{d} \in \mathbb{R}^{2}$ means learnable parameter, and $b_d \in \mathbb{R}^{2}$ is the bias.

Finally, the disease prediction result can be generated using the following equations:
\begin{equation}
    \hat{y} = Softmax(W_{y1} h_{st} + W_{y2} h_{dy} + W_{y3} h_{var} + b_{y})  \vspace{1.0ex} 
\end{equation}
where $w_{y1} \in \mathbb{R}^{d \times d}$, $w_{y2} \in \mathbb{R}^{d \times ((q_{a}+q_{s}+q_{l})*c)}$, $w_{y3} \in \mathbb{R}^{d \times (m-1)}$ are the learnable parameters, and $b_{y} \in \mathbb{R}^{d}$ is the bias.

\subsection{Loss Function and Optimizer}
In this paper, our loss function is cross entropy. This loss function is formalized as:

\begin{equation}
	\mathcal{L} = -\frac{1}{N}\sum_{i=1}^N\sum_{j=1}^D y_{i,j}\log(p_{i,j})  \vspace{1.0ex} 
\end{equation}
where $N$ is the number of patients. $y_{i, j}$ means whether patient $i$ belongs to label $j$, i.e, $y_{i,j}=1$ or $0$. $p_{i,j}$ denotes the prediction probability of the model for patient $i$ belonging to the label $j$. 

We adopt adaptive moment estimation (Adam) as the optimizer in the training process of MPRE. Adam is an optimization algorithm that utilizes first-order gradients to optimize the stochastic objective function~\cite{3.6}. 

\section{Experiments}


In this section, we empirically study the performance of the MPRE. First, we describe the two real-world public datasets. Second, we introduce the baseline methods. Then, we describe the evaluation metrics of experiments. After that, we give the experiment environment. Finally, we discuss the results of the experiments (i.e., performance analysis, ablation study, analysis of symlets, and case study).

\begin{table}[]\centering
\renewcommand{\arraystretch}{1.5}
\caption{ICD-9 Code for Circulatory Disease}
\label{icd}
\begin{tabular}{ll}
\hline
\multicolumn{1}{c}{ICD-9 Code} & \multicolumn{1}{c}{Label} \\ \hline
          393 - 398                     &       chronic rheumatic heart disease                    \\
          401 - 405                     &         hypertensive disease                  \\
          410 - 414                     &         ischemic heart disease                  \\
          415 - 417                     & diseases of pulmonary circulation \\
          420 - 429                     &         other forms of heart disease                  \\
          430 - 438                     &          cerebrovascular disease                 \\
        440 - 449                    &              diseases of arteries, arterioles, and capillaries             \\
          451 - 459                     &        diseases of veins and lymphatics              \\ \hline
\end{tabular}
\end{table}

\subsection{Dataset}
SCRIPT CarpeDiem Dataset~\cite{data1}: The dataset includes 12,495 visit records from 585 patients between June 2018 to March 2022. We used this dataset for respiratory disease prediction. Specifically, 190 patients had COVID-19, 50 had respiratory viral pneumonia, 252 had bacterial pneumonia, and 93 had respiratory failure. 

Health Facts Database~\cite{data2}: The database contains 101,767 visit records for 71,518 patients between 1999 and 2008. In this paper, we predict whether diabetic patients will suffer from circulatory disease in the future. Therefore, the 30,389 visit records for 26,807 patients are selected to form the dataset. We define the labels based on the World Health Organization's ICD-9 codes categories for circulatory diseases (i.e., Table~\ref{icd})~\cite{4.1, n6}. Specifically, 181 patients had chronic rheumatic heart disease, 1469 had hypertensive disease, 9526 had ischemic heart disease, 504 had diseases of pulmonary circulation, 8558 had other forms of heart disease, 4157 had cerebrovascular disease, 1248 had diseases of arteries, arterioles, and capillaries, 1164 had diseases of veins and lymphatics.

\begin{table*}[!ht]
\renewcommand{\arraystretch}{1.5}
    \caption{Average Performances of Proposed MPRE and Baseline Methods}
    \label{result}
    \begin{center}
    \begin{tabular}{m{2.0cm}<{\centering}|m{2.0cm}<{\centering}|m{2.0cm}<{\centering}|m{2.0cm}<{\centering}|m{2.0cm}<{\centering}}
    \hline
    \multirow{2}{*}{\textbf{Model}}&\multicolumn{2}{c|}{\textbf{SCRIPT CarpeDiem Dataset}}&\multicolumn{2}{c}{\textbf{Health Facts Database}}\\
        \cline{2-5} 
        & \textbf{AUROC}& \textbf{AUPRC}& \textbf{AUROC}& \textbf{AUPRC} \\ \hline
        GRU~\cite{4.2} & 0.7528  & 0.6405   & 0.7377  & 0.6234  \\
        TCN~\cite{4.3} & 0.8009  & 0.6751   & 0.7209  & 0.6325  \\
        RETAIN~\cite{99.1} & 0.7612  & 0.6524   & 0.7431  & 0.6190  \\
        T-LSTM~\cite{99.2} & 0.7338  & 0.6274   & 0.7014  & 0.5978  \\
        Dipole~\cite{99.3} & 0.8324   & 0.7428  & 0.7398  & 0.6284  \\
        SAnD~\cite{99.4} & 0.7482  & 0.6316  & 0.7263  & 0.6271  \\
        AdaCare~\cite{99.5} & 0.7641  & 0.6449   & 0.7106  & 0.6092  \\
        StageNet~\cite{4.5} & 0.8183  & 0.7232    & 0.7326  & 0.6297 \\
        ConCare~\cite{n16} & 0.8425 &  0.7531 &  0.7573  & 0.6507
          \\ 
        \textbf{Ours} & \textbf{0.8948}  & \textbf{0.8270}  & \textbf{0.8675}  & \textbf{0.7209}   \\ \hline
    \end{tabular}
    \end{center}
\end{table*}

\subsection{Baselines}
We compare the proposed MPRE with the following baseline models.
\begin{itemize}
    \item [\textbullet] GRU~\cite{4.2} is the standard model for time series data. The embedded patients' features are inputted into the GRU for disease prediction. 
    \item [\textbullet] TCN~\cite{4.3} used the 1D convolutional layer to capture long-term dependencies in time series and also improved the performance of the model by residual structure and dilated convolution.
    \item [\textbullet] RETAIN~\cite{99.1} proposed the attention-based RNN network for learning the weight of visit. This model focused on the recent patient's visit record.
    \item [\textbullet] T-LSTM~\cite{99.2} improved the LSTM model by adopting the time-aware mechanism. This model learned the subspace of cell memory to avoid time decay, and it can be used for irregular time intervals.
    \item [\textbullet] Dipole~\cite{99.3} designed bidirectional RNN with the attention mechanism to analyze the patients' visit records. It can use the attention weight to compare the importance of each visit.
    \item [\textbullet] SAnD~\cite{99.4} introduced a Transformer-based model that utilized masked self-attention to effectively model clinical time series data. This approach allowed for the integration of information from multiple visit records. 
    \item [\textbullet] AdaCare~\cite{99.5} adopted the 1D dilated convolution network to capture the different variations of biomarkers at different scales and followed by the GRU model for disease prediction.
    \item [\textbullet] StageNet~\cite{4.5} used the stage-aware long short-term memory to extract the progression of health status and utilized the convolutional layer to capture the underlying progression patterns. 
    \item[\textbullet] ConCare~\cite{n16} embedded each feature separately and adopted the time-aware attention mechanism and information decay function for disease prediction.
\end{itemize}

\subsection{Evaluation Metrics}
To evaluate the disease prediction task, we adopt the area under the receiver operating characteristic curve (AUROC) and precision-recall curve (AUPRC) as evaluation metrics to compare the performance of the proposed MPRE and baseline methods. Note that AUPRC is the primary evaluation metric for the imbalanced dataset~\cite{4.6}.

\begin{figure*}[!t]
\centerline{\includegraphics[width=0.81\textwidth,height=0.57
\textwidth]{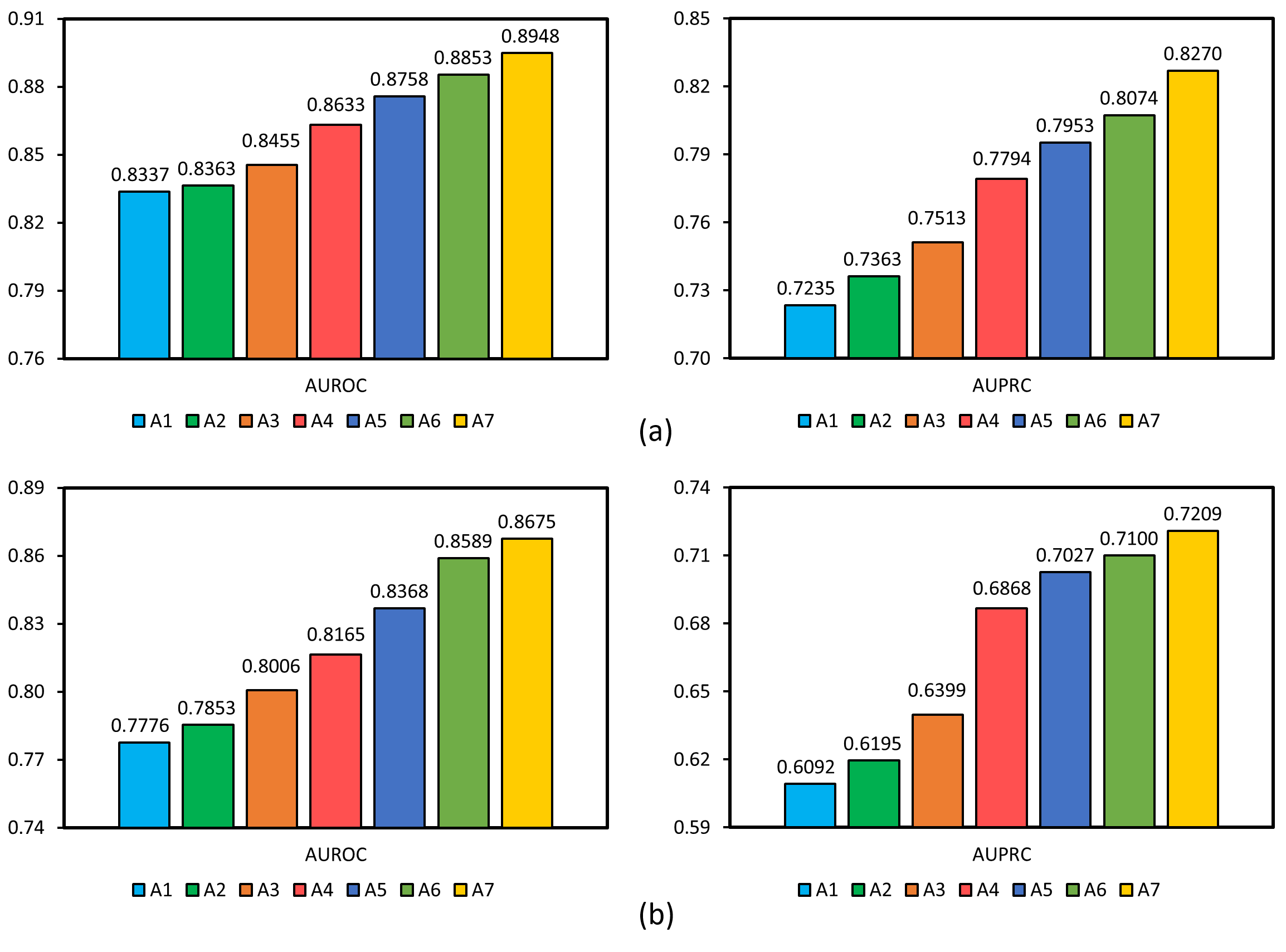}}
\caption{The performance of ablation studies in terms of AUROC and AUPRC. (a) shows the average performance on SCRIPT CarpeDiem Dataset. (b) presents the average performance on Health Facts Database.}
\label{abs}
\end{figure*}

\subsection{Experiment Environment}
We implement MPRE and baseline methods with Python 3.8 and PyTorch framework\footnote{https://pytorch.org/}. The experiments are conducted on the machine equipped with Nvidia GPU RTX A6000. We report the optimized parameters as follows. The learning rate is set to $10^{-4}$, and the batch size is 64. We adopt Tanh as the activation function. In MPRE, the symlets-18 wavelet is applied to the SCRIPT CarpeDiem Dataset, while the symlets-14 wavelet is utilized for the Health Facts Database. For 2D MEN, the kernel size is set to 1. To extract the adjacent correlation, we are not using the dilated rate. To extract the short-term correlation, the dilated rate is set to 1. To extract the long-term correlation, the dilated rate is set to 3. For the sake of fairness, we implement all baselines on the same platform and parameters. We adopt the 10-fold cross-validation and report the average performance of MPRE and baseline methods.

\subsection{Performance Analysis}

The average performances of MPRE and nine baseline methods in terms of AUROC and AUPRC are shown in Table~\ref{result}. We can see that our proposed MPRE outperforms all baseline methods, achieving AUROC of 0.8948 and AUPRC of 0.8270 on the SCRIPT CarpeDiem Dataset, as well as AUROC of 0.8675 and AUPRC of 0.7209 on the Health Facts Database.

In SCRIPT CarpeDiem Dataset, we find that ConCare achieves better performance than other baseline methods, i.e., AUROC of 0.8425 and AUPRC of 0.7531. This is because ConCare embedded the features separately, which better learned the information of each feature. In addition, ConCare adopted the time decay function to better learn the patients' disease progression. Compared with ConCare, MPRE improves AUROC by 5.84\% and 8.94\% by AUPRC. In addition, we observe that T-LSTM achieves worse results. One possible explanation for this result is that the patient information does not exhibit a monotonic decrease with increasing time intervals.

In Health Facts Database, we observe that ConCare also achieves the highest AUROC among all the baseline methods, with AUROC of 0.7573, AUPRC of 0.6507. Compared with ConCare, MPRE improves AUROC by 12.70\%, and a 9.74\% improvement in AUPRC. Besides, we find that the T-LSTM model still has the worst performance among the baseline methods.  

We give an explanation for why the MPRE is capable of achieving superior performance by highlighting three key factors. First, FTM transforms the dynamic feature information of limited visit records into the time-frequency domain to enhance the feature representations. Second, the 2D MEN is capable of capturing the correlation between trends and variations simultaneously. Finally, the FODAM is used to adaptively identify the contribution of differences in variations to the disease diagnosis.

\subsection{Ablation Study}
To evaluate the effectiveness of our proposed framework, we conduct ablation studies on MPRE. Table~\ref{table_config} displays the configuration for the ablation studies, where a “$\checkmark$” signifies the inclusion of the corresponding module in the experiments and a “$\times$” indicates its exclusion. The first three columns list the modules used in the MPRE, while the last two columns display the trend and variation obtained from the FTM.
 
\begin{itemize}
    \item A1 solely employs FTM to extract trend information in the experiment, aiming to evaluate the performance of the model using only trend information.
    \item A2 exclusively utilizes FTM to obtain variation for disease prediction, aiming to present the model's performance using only variations.
    \item A3 employs FTM to obtain variation and FODAM to calculate the contributions of differences in adjacent variation to disease diagnosis, aiming to assess the importance of utilizing variation and differences.
    \item A4 utilizes trend and variation from FTM, aiming to explore the importance of simultaneously using trend and variation information
    \item A5 adopts trend and variation information from FTM and incorporates FODAM, aiming to evaluate the model's performance when exploiting trends, variations, and differences simultaneously.
    \item A6 employs 2D MEN and FTM from MPRE, aiming to investigate the importance of capturing correlations between trend and variation information.
    \item A7 is the complete MPRE.
\end{itemize}

\begin{figure*}[!t]
\centerline{\includegraphics[width=0.9914\textwidth,height=0.38
\textwidth]{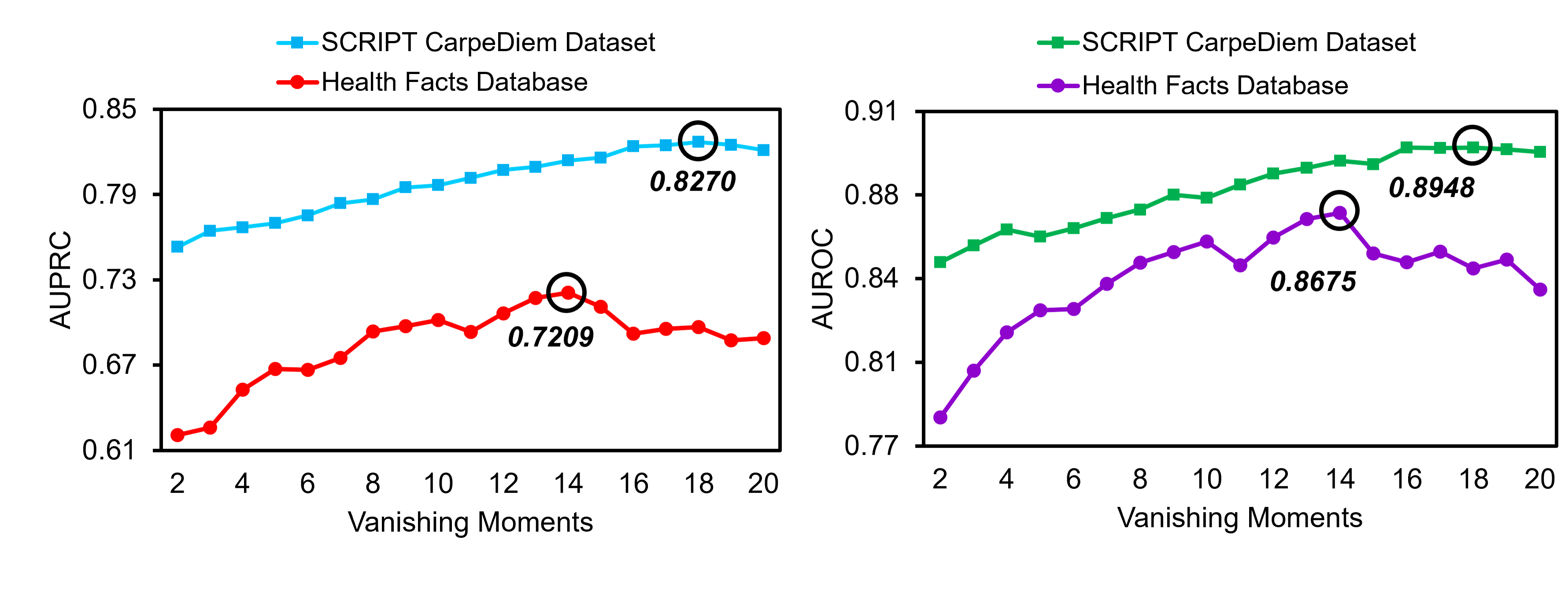}}
\caption{The performance of symlets with different vanishing moments in terms of AUPRC and AUROC.}
\label{para}
\end{figure*}

\begin{table}[]\centering
\renewcommand{\arraystretch}{1.5}
\caption{Configurations for Ablation Studies}
\begin{tabular}{cccccc}
\hline
Configurations & FTM & 2D MEN & FODAM & Trend & Variation \\ \hline
A1             & $\checkmark$    &    $\times$    &  $\times$  &   $\checkmark$  & $\times$  \\
A2             &  $\checkmark$   &     $\times$   &  $\times$  &  $\times$    & $\checkmark$ \\
A3             & $\checkmark$    &   $\times$     &  $\checkmark$ &    $\times$       &  $\checkmark$  \\
A4             &  $\checkmark$   &    $\times$    &  $\times$  & $\checkmark$  &  $\checkmark$  \\
A5             & $\checkmark$    &    $\times$    &   $\checkmark$ &  $\checkmark$  & $\checkmark$  \\
A6             &  $\checkmark$   &   $\checkmark$     &  $\times$  &   $\checkmark$    &  $\checkmark$ \\ 
A7             &  $\checkmark$   &   $\checkmark$     &  $\checkmark$  &   $\checkmark$    &  $\checkmark$ \\ \hline
\end{tabular}
\label{table_config}
\end{table}

\begin{table*}[]
\centering
\renewcommand{\arraystretch}{1.4}
    \caption{Top 5 Correlations between Trend and Variation in Dynamic Features among Four Respiratory Diseases}
    \label{casestudy}
\begin{tabular}{c|c}
\hline
Types of Respiratory Disease                 & Top 5 Correlations between the Trend and Variation \\ \hline
\multirow{5}{*}{Bacterial Pneumonia}         & diastolic blood pressure (0.85)                   \\
                                             & hemoglobin (0.83)                                 \\
                                             & mean arterial pressure (0.81)                     \\
                                             & systolic blood pressure (0.81)                    \\
                                             & heart rate (0.81)                                 \\ \hline
\multirow{5}{*}{Respiratory Viral Pneumonia} & Platelets (0.80)                                  \\
                                             & blood pressure (0.73)                             \\
                                             & diastolic blood pressure (0.73)                   \\
                                             & respiratory rate (0.73)                           \\
                                             & heart rate (0.73)                                 \\ \hline
\multirow{5}{*}{COVID-19}                    & lymphocytes (0.87)                                \\
                                             & peep changes (0.74)                               \\
                                             & fio2 (0.72)                                       \\
                                             & peep (0.70)                                       \\
                                             & respiratory rate changes (0.68)                   \\ \hline
\multirow{5}{*}{Respiratory Failure}         & bicarbonate (0.79)                                \\
                                             & heart rate (0.59)                                 \\
                                             & urine output (0.58)                               \\
                                             & platelets (0.55)                                  \\
                                             & diastolic blood pressure (0.54)                   \\ \hline
\end{tabular}
\end{table*}

The results of our ablation studies are presented in Figure~\ref{abs}. We can observe that the utilization of variation information (A2) yields superior model performance as compared to solely utilizing trend information (A1). Compared to relying solely on variation information (A2), the utilization of variation information and differences (A3) can further improve the model performance. Compared to A1, A2, and A3, we find that the model's performance improves when both trend and variation information is used (A4). Our empirical findings suggest that the performance of the model can be further improved by employing FODAM in conjunction with the A4-based configurations, which we refer to as the configurations of A5. In A6, we adopt the 2D MEN to extract the correlation between the trend and variation information. In other words, we capture the trend information with significant variations. Therefore, in comparison to A1, A2, and A4, we observe a significant improvement in the model performance with the implementation of the A6 configuration. A7 means the complete MPRE, which is superior to the experimental results of A1 to A6. Based on the results of these ablation studies, we can conclude that the design of the proposed MPRE is reasonable and effective in capturing the critical features for disease prediction.

\subsection{Analysis of Symlets}

We conduct experiments to evaluate the performance of various types of symlets, which differ in the number of vanishing moments they possess. A larger value of vanishing moments means that the wavelet function is able to eliminate higher-order polynomial signals, thus improving the smoothness and compression ratio of the signal. In this paper, we try 19 candidate vanishing moments values, i.e., symlets-2 to symlets-20. In all datasets, we implement our method under these 19 candidate vanishing moments and report the average performance in terms of AUPRC and AUROC.

In Figure~\ref{para}, we show the performance of different types of symlets on the SCRIPT CarpeDiem dataset. We find that the optimal vanishing moment value of symlets is 18. While the symlets-16 and symlets-18 have the same AUROC values, symlets-18 outperforms symlets-16 in terms of AUPRC, with values of 0.8270 and 0.8238, respectively. The performance of Health Facts Database is also shown in Figure~\ref{para}. We can observe that the optimal vanishing moment value is 14. Based on the results from the two datasets, we can conclude that both excessively high or low values of vanishing moments are not desirable. Excessively high vanishing moments may lead to insufficient capture of details and subtle variations, while excessively low vanishing moments may add redundant information.

\begin{figure*}[!t]
\centerline{\includegraphics[width=0.997\textwidth,height=0.35
\textwidth]{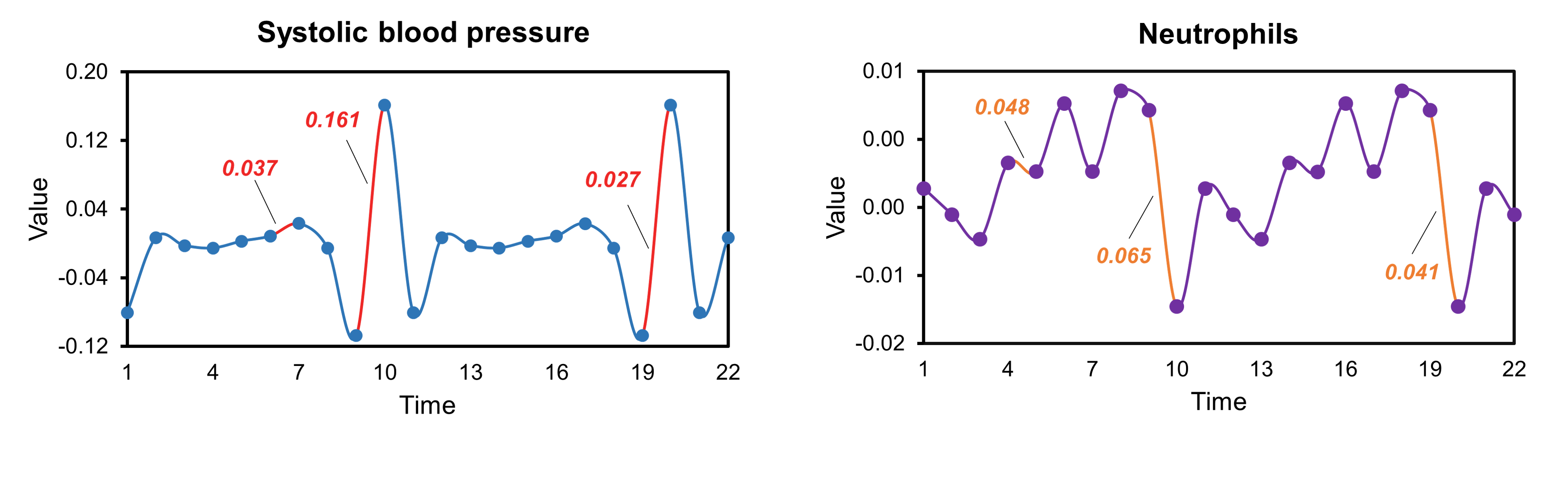}}
\caption{The attention scores for differences of adjacent variation in two dynamic features, i.e., systolic blood pressure and neutrophils.}
\label{secondcasestudy}
\end{figure*}

\subsection{Case Study}
In this paper, we adopt the 2D MEN to capture the correlation between the trend and variation. To quantitatively support the existence of the correlation between the trend and variation, we conduct experiments to compute the Pearson correlation. We randomly select four patients from the SCRIPT CarpeDiem Dataset as examples. The selected patients are from four different types of respiratory disease, respectively, i.e., bacterial pneumonia, respiratory viral pneumonia, COVID-19, and respiratory failure. Table~\ref{casestudy} shows the Top 5 correlations between the trend and variation of each type of respiratory disease. The larger the value, the stronger the correlation. We can observe that the strongest correlations between trends and variations in bacterial pneumonia were diastolic blood pressure, hemoglobin, mean arterial pressure, systolic blood pressure, and heart rate. As for respiratory viral pneumonia, the most prominent correlations are found in platelets, blood pressure, diastolic blood pressure, respiratory rate, and heart rate. In COVID-19, lymphocytes, peep changes, fio2, peep, and respiratory rate changes display the strongest correlations between the trend and variation. Finally, in the case of respiratory failure, the strongest correlations are bicarbonate, heart rate, urine output, platelets, and diastolic blood pressure. By capturing the correlation between the trend and variation, our model provides correct predictions for all these four patients.

We also randomly select two dynamic features from the SCRIPT CarpeDiem Dataset and show their results from FODAM (i.e., the attention scores on differences of adjacent variations). Based on Figure~\ref{secondcasestudy}, we can observe that the attention scores reach their pinnacle when the patient undergoes the initial substantial rise or fall in contiguous fluctuations (i.e., 0.161 for the primary notable increase in systolic blood pressure and 0.065 for the primary notable decrease in neutrophils). This indicates that this is an important shift in patient health status. For the subsequent significant increase or decrease, the attention scores exhibit a decline and are lower than the attention scores of slight rise or fall (i.e., the score of 0.037 and 0.027 for slight improvement and the second significant improvement in systolic blood pressure, respectively. The score of 0.048 and 0.041 for a slight decrease and the second significant decrease in neutrophils, respectively). This result suggests that physicians should pay attention to early changes in the patient's health status, which aligns with the clinical diagnosis guidance~\cite{n17,n18, bohr2020rise}.

\section{Conclusion}

In this paper, we propose Multi-perspective Patient Representation Extractor called MPRE for disease prediction. Specifically, the proposed FTM obtains trend and variation information from dynamic features. FTM can enhance the dynamic feature representations by time-frequency transformation, which addresses the challenge of data sparsity.
Then, the proposed 2D MEN is utilized to capture the adjacent, short and long-term correlation between the trend and variation. In addition, we propose FODAM to compute the contributions of differences in dynamic feature variations to disease diagnosis. We evaluate the MPRE and state-of-the-art baseline methods on two real-world public datasets, namely "SCRIPT CarpeDiem Dataset" and "Health Facts Database". The experiment results show that our MPRE method outperforms the baseline methods.

\section{Acknowledge}
This work was supported in part by the Macao Polytechnic University – Research on Representation Learning in Decision Support for Medical Diagnosis (RP/FCA-11/2022). 


\bibliographystyle{IEEEtran} 
\IEEEtriggeratref{28}
\bibliography{reference}

\end{document}